\documentclass{article}

\usepackage[preprint]{neurips_2023}

\usepackage[utf8]{inputenc} 
\usepackage[T1]{fontenc}    
\usepackage{hyperref}       
\usepackage{url}            
\usepackage{booktabs}       
\usepackage{amsfonts}       
\usepackage{nicefrac}       
\usepackage{microtype}      
\usepackage{xcolor}         

\usepackage{float}

\usepackage{wrapfig}

\usepackage{graphicx}
\usepackage{amsmath}

\usepackage{listings}
\definecolor{codegreen}{rgb}{0,0.6,0}
\definecolor{codegray}{rgb}{0.5,0.5,0.5}
\definecolor{codepurple}{rgb}{0.58,0,0.82}
\definecolor{backcolour}{rgb}{0.95,0.95,0.92}
\newcommand{\Secref}[1]{Section~\ref{#1}}
\def\Figref#1{Figure~\ref{#1}}

\usepackage{xspace}

\newcommand{\ie}{\emph{i.e.},\xspace}

\title{94\% on CIFAR-10 in 3.29 Seconds on a Single GPU}

\author{
Keller Jordan \\
\texttt{kjordan4077@gmail.com}
}

\begin{document}

\maketitle

\vspace{-2mm}
\begin{abstract}
\vspace{-2mm}
CIFAR-10 is among the most widely used datasets in machine learning, facilitating thousands of research projects per year.
To accelerate research and reduce the cost of experiments, we introduce training methods for CIFAR-10 which reach 94\% accuracy in 3.29 seconds, 95\% in 10.4 seconds, and 96\% in 46.3 seconds, when run on a single NVIDIA A100 GPU.
As one factor contributing to these training speeds, we propose a derandomized variant of horizontal flipping augmentation, which we show improves over the standard method in every case where flipping is beneficial over no flipping at all.
Our code is released at \url{https://github.com/KellerJordan/cifar10-airbench}.
\end{abstract}

\section{Introduction}

CIFAR-10~\citep{cifar100} is one of the most popular datasets in machine learning, facilitating thousands of research projects per year\footnote{\url{https://paperswithcode.com/datasets}}. 
Research can be accelerated and the cost of experiments reduced if the speed at which it is possible to train neural networks on CIFAR-10 is improved.
In this paper we introduce a training method which reaches 94\% accuracy in 3.29 seconds on a single NVIDIA A100 GPU, which is a $1.9\times$ improvement over the prior state-of-the-art~\citep{hlbCIFAR10}.
To support scenarios where higher performance is needed, we additionally develop methods targeting 95\% and 96\% accuracy. We release the following methods in total.
\begin{enumerate}
    \item \href{https://github.com/KellerJordan/cifar10-airbench/blob/04512094b7341d95ed02f697c5f5db404556137e/airbench94_compiled.py}{\texttt{airbench94\_compiled.py}}: \textbf{94.01\% accuracy in 3.29 seconds} ($3.6 \times 10^{14}$ FLOPs).
    \item \href{https://github.com/KellerJordan/cifar10-airbench/blob/04512094b7341d95ed02f697c5f5db404556137e/airbench94.py}{\texttt{airbench94.py}}: 94.01\% accuracy in 3.83 seconds ($3.6 \times 10^{14}$ FLOPs).
    \item \href{https://github.com/KellerJordan/cifar10-airbench/blob/04512094b7341d95ed02f697c5f5db404556137e/airbench95.py}{\texttt{airbench95.py}}: 95.01\% accuracy in 10.4 seconds ($1.4 \times 10^{15}$ FLOPs).
    \item \href{https://github.com/KellerJordan/cifar10-airbench/blob/04512094b7341d95ed02f697c5f5db404556137e/airbench96.py}{\texttt{airbench96.py}}: 96.05\% accuracy in 46.3 seconds ($7.2 \times 10^{15}$ FLOPs).
\end{enumerate}
All runtimes are measured on a single NVIDIA A100.
We note that the first two scripts are mathematically equivalent (\ie yield the same distribution of trained networks), and differ only in that the first uses \texttt{torch.compile} to improve GPU utilization. It is intended for experiments where many networks are trained at once in order to amortize the one-time compilation cost.
The non-compiled \texttt{airbench94} variant can be easily installed and run using the following command.
\begin{lstlisting}[language=Python]
pip install airbench
python -c "import airbench as ab; ab.warmup94(); ab.train94()"
\end{lstlisting}

One motivation for the development of these training methods is that they can accelerate the experimental iteration time of researchers working on compatible projects involving CIFAR-10. Another motivation is that they can decrease the cost of projects involving a massive number of trained networks. One example of such a project is \citet{ilyas2022datamodels}, a study on data attribution which used 3 million trained networks to demonstrate that the outputs of a trained neural network on a given test input follow an approximately linear function of the vector of binary choices of which examples the model was trained on. Another example is \citet{jordan2023calibrated}, a study on training variance which used 180 thousand trained networks to show that standard trainings have little variance in performance on their test-distributions. These studies were based on trainings which reach 93\% in 34 A100-seconds and 94.4\% in 72 A100-seconds, respectively. The training methods we introduce in this paper make it possible to replicate these studies, or conduct similar ones, with fewer computational resources.

Fast training also enables the rapid accumulation of statistical significance for subtle hyperparameter comparisons. For example, if changing a given hyperparameter subtly improves mean CIFAR-10 accuracy by 0.02\% compared to a baseline, then (assuming a typical 0.14\% standard deviation between runs~\citep{jordan2023calibrated}) we will need on average $N = 133$ runs of training to confirm the improvement at a statistical significance of $p = 0.05$.
For a standard 5-minute ResNet-18 training this will take 11.1 GPU-hours; \texttt{airbench94} shrinks this to a more convenient time of 7.3 minutes.

\begin{figure}
    \centering
    \includegraphics[width=0.49\textwidth]{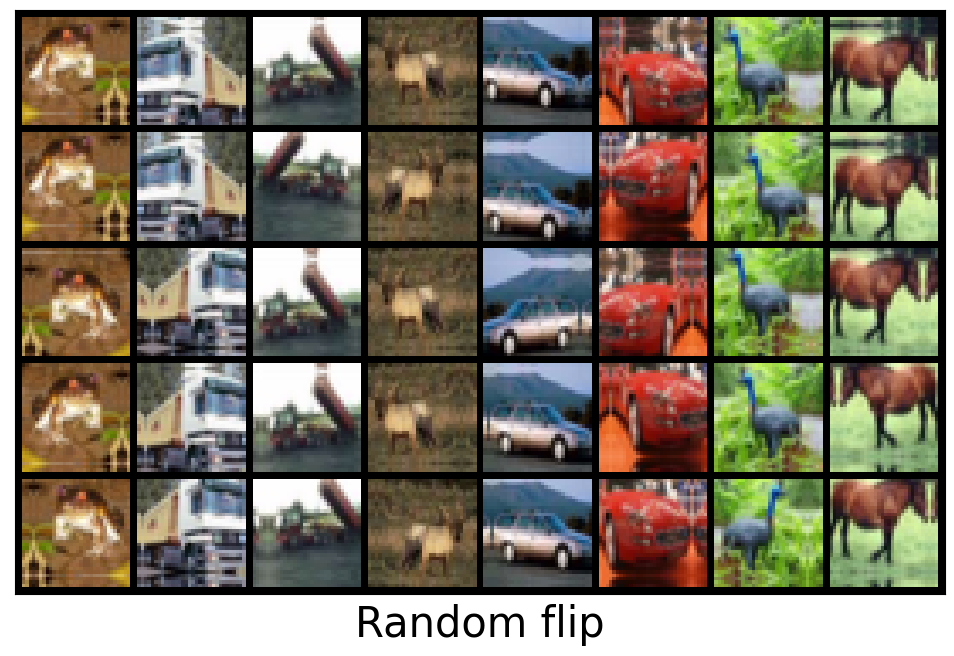}
    \includegraphics[width=0.49\textwidth]{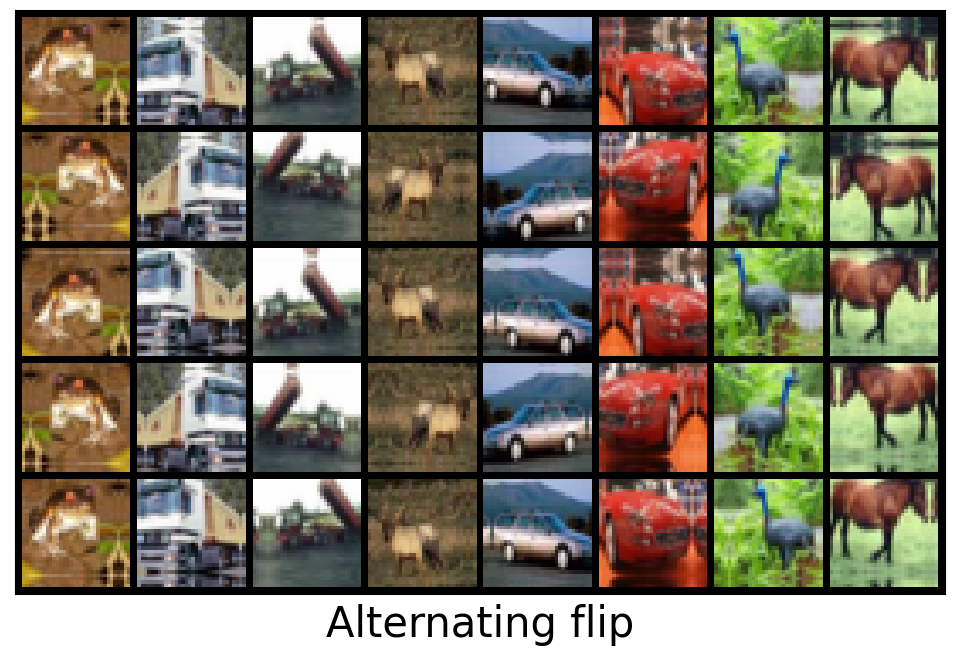}
    \caption{\small
    \textbf{Alternating flip.}
    In computer vision we typically train neural networks using random horizontal flipping augmentation, which flips each image with 50\% probability per epoch.
    This results in some images being redundantly flipped the same way for many epochs in a row.
    We propose~(\Secref{sec:altflip}) to flip images in a deterministically alternating manner after the first epoch, avoiding this redundancy and speeding up training.
    }
    \label{fig:viz_altflip}
    \vspace{-2mm}
\end{figure}

Our work builds on prior training speed projects. We utilize a modified version of the network, initialization, and optimizer from \citet{hlbCIFAR10}, as well as the optimization tricks and frozen patch-whitening layer from \citet{paged2019resnet,hlbCIFAR10}. The final $\sim$10\% of our speedup over prior work is obtained from a novel improvement to standard horizontal flipping augmentation~(\Figref{fig:viz_altflip}, \Secref{sec:altflip}, \Secref{sec:experiments_fliplr}).




\vspace{-2mm}
\section{Background}
\vspace{-2mm}

Our objective is to develop a training method which reaches 94\% accuracy on the CIFAR-10 test-set in the shortest possible amount of time. Timing begins when the method is first given access to training data, and ends when it produces test-set predictions. The method is considered valid if its mean accuracy over repeated runs is at least 94\%.

We chose the goal of 94\% accuracy because this was the target used by the CIFAR-10 track of the 2017-2020 Stanford DAWNBench training speed competition~\citep{coleman2017dawnbench}, as well as more recent work~\citep{hlbCIFAR10}. The final winning DAWNBench submission reached 94\% in 10 seconds on 8 V100s~\citep{applecifar2019} ($\approx 32$ A100-seconds), using a modified version of \citet{paged2019resnet}, which itself runs in 26 V100-seconds ($\approx 10.4$ A100-seconds). The prior state-of-the-art is \citet{hlbCIFAR10} which attains 94\% in 6.3 A100-seconds. As another motivation for the goal, 94\% is the level of human accuracy reported by \citet{karpathy2011lessons}.

We note the following consequences of how the method is timed. First, it is permitted for the program to begin by executing a run using dummy data in order to ``warm up'' the GPU, since timing begins when the training data is first accessed. This is helpful because otherwise the first run of training is typically a bit slower. Additionally, arbitrary test-time augmentation (TTA) is permitted. TTA improves the performance of a trained network by running it on multiple augmented views of each test input. Prior works~\citep{paged2019resnet,applecifar2019,hlbCIFAR10} use horizontal flipping TTA; we use horizontal flipping and two extra crops. Without any TTA our three training methods attain 93.2\%, 94.4\%, and 95.6\% mean accuracy respectively.

The CIFAR-10 dataset contains 60,000 32x32 color images, each labeled as one of ten classes. It is divided into a training set of 50,000 images and a validation set of 10,000 images.
As a matter of historical interest, we note that in 2011 the state-of-the-art accuracy on CIFAR-10 was 80.5\%~\citep{cirecsan2011high}, using a training method which consumes $26\times$ \textit{more} FLOPs than \texttt{airbench94}.
Therefore, the progression from 80.5\% in 2011 to the 94\% accuracy of \texttt{airbench94} can be attributed entirely to algorithmic progress rather than compute scaling.

\vspace{-1mm}
\section{Methods}
\vspace{-1mm}


\subsection{Network architecture and baseline training}
\label{sec:network}
\vspace{-1mm}

We train a convolutional network with a total of 1.97 million parameters, following \citet{hlbCIFAR10} with a few small changes. It contains seven convolutions with the latter six being divided into three blocks of two. The precise architecture is given as simple PyTorch code in \Secref{sec:arch}; in this section we offer some comments on the main design choices.


The network is VGG~\citep{simonyan2014very}-like in the sense that its main body is composed entirely of 3x3 convolutions and 2x2 max-pooling layers, alongside BatchNorm~\citep{ioffe2015batch} layers and activations. Following \citet{hlbCIFAR10} the first layer is a 2x2 convolution with no padding, causing the shape of the internal feature maps to be 31x31 → 15x15 → 7x7 → 3x3 rather than the more typical 32x32 → 16x16 → 8x8 → 4x4, resulting in a slightly more favorable tradeoff between throughput and performance. We use GELU~\citep{hendrycks2016gaussian} activations.

Following \citet{paged2019resnet,hlbCIFAR10}, we disable the biases of convolutional and linear layers, and disable the affine scale parameters of BatchNorm layers. The output of the final linear layer is scaled down by a constant factor of 1/9.
Relative to \citet{hlbCIFAR10}, our network architecture differs only in that we decrease the number of output channels in the third block from 512 to 256, and we add learnable biases to the first convolution.

As our baseline, we train using Nesterov SGD at batch size 1024, with a label smoothing rate of 0.2. We use a triangular learning rate schedule which starts at $0.2\times$ the maximum rate, reaches the maximum at 20\% of the way through training, and then decreases to zero. For data augmentation we use random horizontal flipping alongside 2-pixel random translation. For translation we use reflection padding~\citep{zagoruyko2016wide} which we found to be better than zero-padding. Note that what we call 2-pixel random translation is equivalent to padding with 2 pixels and then taking a random 32x32 crop. During evaluation we use horizontal flipping test-time augmentation, where the network is run on both a given test image and its mirror and inferences are made based on the average of the two outputs. With optimized choices of learning rate, momentum, and weight decay, this baseline training configuration yields 94\% mean accuracy in 45 epochs taking 18.3 A100-seconds.

\vspace{-1mm}
\subsection{Frozen patch-whitening initialization}
\label{sec:whiten}
\vspace{-1mm}

Following \citet{paged2019resnet,hlbCIFAR10} we initialize the first convolutional layer as a patch-whitening transformation. The layer is a 2x2 convolution with 24 channels. Following \citet{hlbCIFAR10} the first 12 filters are initialized as the eigenvectors of the covariance matrix of 2x2 patches across the training distribution, so that their outputs have identity covariance matrix. The second 12
\begin{wrapfigure}{r}{0.45\textwidth}
\vspace{-3mm}
\begin{center}
    \includegraphics[width=0.44\textwidth]{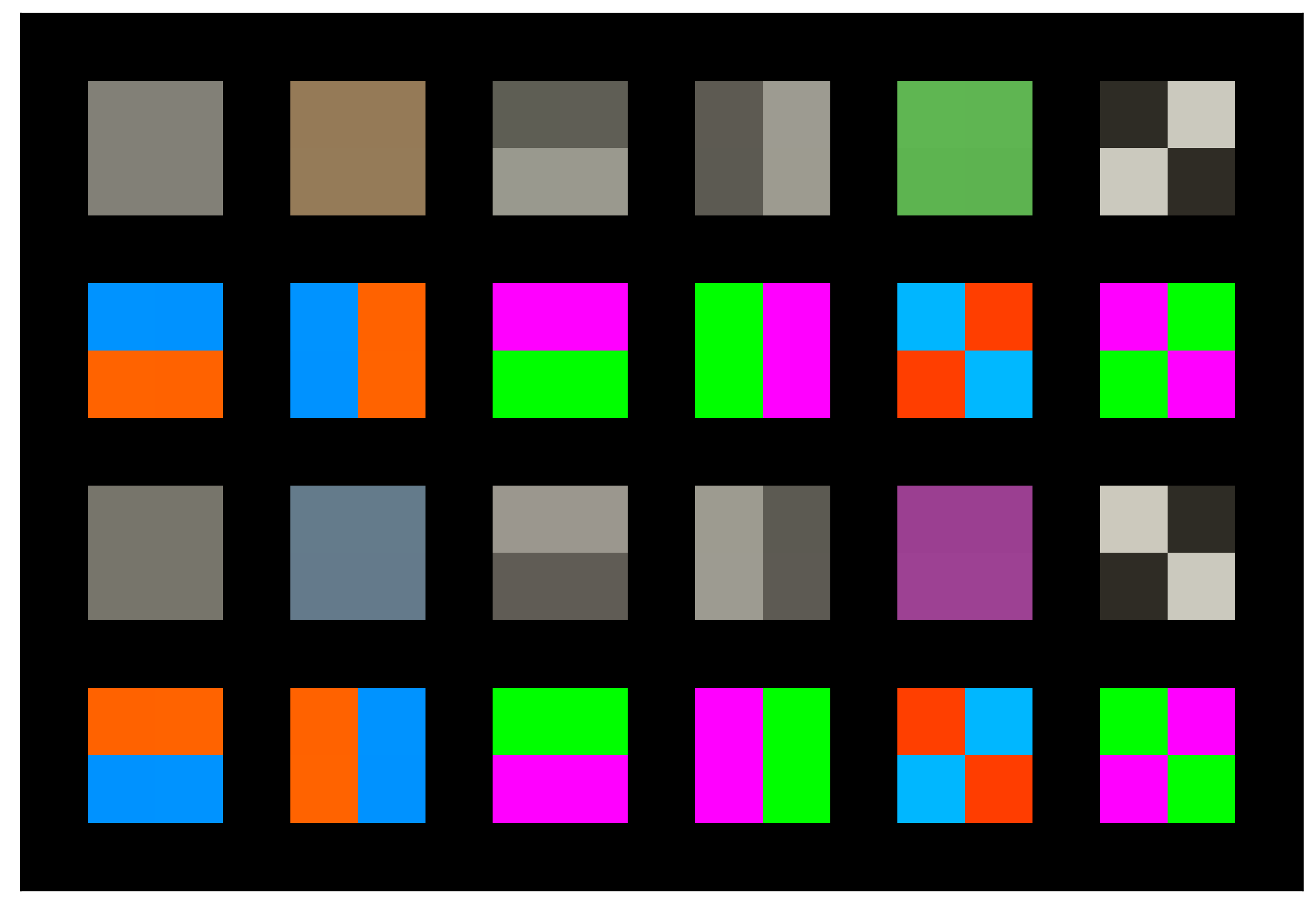}
    \caption{\small The first layer's weights after whitening initialization~\citep{hlbCIFAR10,paged2019resnet}}
    \label{fig:whiten}
\end{center}
\vspace{-5mm}
\end{wrapfigure}
filters are
initialized as the negation of the first 12, so that input information is preserved through the activation which follows.
\Figref{fig:whiten} shows the result.  We do not update this layer's weights during training.

Departing from \citet{hlbCIFAR10}, we add learnable biases to this layer, yielding a small performance boost. The biases are trained for 3 epochs, after which we disable their gradient to increase backward-pass throughput, which improves training speed without reducing accuracy. We also obtain a slight performance boost relative to \citet{hlbCIFAR10} by reducing the constant added to the eigenvalues during calculation of the patch-whitening initialization for the purpose of preventing numerical issues in the case of a singular patch-covariance matrix.

Patch-whitening initialization is the single most impactful feature. Adding it to the baseline more than doubles training speed so that we reach 94\% accuracy in 21 epochs taking 8.0 A100-seconds.

\vspace{-1mm}
\subsection{Identity initialization}
\vspace{-1mm}
\label{sec:dirac}

\textbf{dirac}: We initialize all convolutions after the first as partial identity transforms. That is, for a convolution with $M$ input channels and $N \geq M$ outputs, we initialize its first $M$ filters to an identity transform of the input, and leave the remaining $N-M$ to their default initialization. In PyTorch code, this amounts to running \texttt{torch.nn.init.dirac\_(w[:w.size(1)])} on the weight \texttt{w} of each convolutional layer. This method partially follows \citet{hlbCIFAR10}, which used a more complicated scheme where the identity weights are mixed in with the original initialization, which we did not find to be more performant. With this feature added, training attains 94\% accuracy in 18 epochs taking 6.8 A100-seconds.

\vspace{-1mm}
\subsection{Optimization tricks}
\vspace{-1mm}
\label{sec:opt}

\textbf{scalebias}: We increase the learning rate for the learnable biases of all BatchNorm layers by a factor of $64\times$, following \citet{paged2019resnet,hlbCIFAR10}. With this feature added, training reaches 94\% in 13.5 epochs taking 5.1 A100-seconds.

\textbf{lookahead}: Following \citet{hlbCIFAR10}, we use Lookahead~\citep{zhang2019lookahead} optimization. We note that Lookahead has also been found effective in prior work on training speed for ResNet-18~\citep{moreau2022benchopt}.
With this feature added, training reaches 94\% in 12.0 epochs taking 4.6 A100-seconds.

\vspace{-1mm}
\subsection{Multi-crop evaluation}
\vspace{-1mm}
\label{sec:multicrop}

\textbf{multicrop}: To generate predictions, we run the trained network on six augmented views of each test image: the unmodified input, a version which is translated up-and-to-the-left by one pixel, a version which is translated down-and-to-the-right by one pixel, and the mirrored versions of all three. Predictions are made using a weighted average of all six outputs, where the two views of the untranslated image are weighted by 0.25 each, and the remaining four views are weighted by 0.125 each. With this feature added, training reaches 94\% in 10.8 epochs taking 4.2 A100-seconds.



We note that multi-crop inference is a classic method for ImageNet~\citep{deng2009imagenet} trainings~\citep{simonyan2014very,szegedy2014going}, where performance improves as the number of evaluated crops is increased, even up to 144 crops~\citep{szegedy2014going}. In our experiments, using more crops does improve performance, but the increase to inference time outweighs the potential training speedup.

\vspace{-1mm}
\subsection{Alternating flip}
\vspace{-1mm}
\label{sec:altflip}

\begin{table}
\centering
\begin{tabular}{ll|l}
    \toprule
    Random reshuffling & Alternating flip & Mean accuracy \\
    \midrule
    No & No & 93.40\% \\
    No & Yes & 93.48\% \\
    Yes & No & 93.92\% \\
    Yes & Yes & 94.01\% \\
    \bottomrule
\end{tabular}
\vspace{3mm}
\caption{\small Training distribution options (\Secref{sec:altflip}). Both random reshuffling (which is standard) and alternating flip (which we propose) reduce training data redundancy and improve performance.}
\label{tab:traindist}
\vspace{-5mm}
\end{table}

To speed up training, we propose a derandomized variant of standard horizontal flipping augmentation, which we motivate as follows. When training neural networks, it is standard practice to organize training into a set of epochs during which every training example is seen exactly once. This differs from the textbook definition of stochastic gradient descent (SGD)~\citep{robbins1951stochastic}, which calls for data to be repeatedly sampled \textit{with-replacement} from the training set, resulting in examples being potentially seen multiple redundant times within a short window of training. The use of randomly ordered epochs of data for training has a different name, being called the \textit{random reshuffling} method in the optimization literature~\citep{gurbuzbalaban2021random,bertsekas2015convex}.
If our training dataset consists of $N$ unique examples, then sampling data with replacement causes every ``epoch'' of $N$ sampled examples to contain only $(1 - (1-1/N)^N)N \approx (1 - 1/e)N \approx 0.632N$ unique examples on average. On the other hand, random reshuffling leads to all $N$ unique examples being seen every epoch. Given that random reshuffling is empirically successful~(Table~\ref{tab:traindist}), we reason that it is beneficial to maximize the number of unique inputs seen per window of training time.

We extend this reasoning to design a new variant of horizontal flipping augmentation, as follows. We first note that standard random horizontal flipping augmentation can be defined as follows.

\begin{lstlisting}[language=Python, caption=Random flip]
import torch
def random_flip(inputs):
    # Applies random flipping to a batch of images
    flip_mask = (torch.rand(len(inputs)) < 0.5).view(-1, 1, 1, 1)
    return torch.where(flip_mask, inputs.flip(-1), inputs)]
\end{lstlisting}
\vspace{-2mm}

If horizontal flipping is the only augmentation used, then there are exactly $2N$ possible unique inputs\footnote{Assuming none of the training inputs are already mirrors of each other.} which may be seen during training. Potentially, every pair of consecutive epochs could contain every unique input. But our main observation is that with standard random horizontal flipping, half of the images will be redundantly flipped the same way during both epochs, so that on average only $1.5N$ unique inputs will be seen.

\textbf{altflip}: To address this, we propose to modify standard random horizontal flipping augmentation as follows.
For the first epoch, we randomly flip 50\% of inputs as usual. 
Then on epochs $\{2, 4, 6, \dots\}$, we flip only those inputs which were not flipped in the first epoch, and on epochs $\{3, 5, 7, \dots\}$, we flip only those inputs which were flipped in the first epoch.
We provide the following implementation which avoids the need for extra memory by using a pseudorandom function to decide the flips.
\begin{lstlisting}[language=Python, caption=Alternating flip]
import torch
import hashlib
def hash_fn(n, seed=42):
    k = n * seed
    return int(hashlib.md5(bytes(str(k), 'utf-8')).hexdigest()[-8:], 16) 
def alternating_flip(inputs, indices, epoch):
    # Applies alternating flipping to a batch of images
    hashed_indices = torch.tensor([hash_fn(i) for i in indices.tolist()])
    flip_mask = ((hashed_indices + epoch) % 2 == 0).view(-1, 1, 1, 1)
    return torch.where(flip_mask, inputs.flip(-1), inputs)
\end{lstlisting}
\vspace{-2mm}

The result is that every pair of consecutive epochs contains all $2N$ unique inputs, as we can see in \Figref{fig:viz_altflip}. We demonstrate the effectiveness of this method across a variety of scenarios in \Secref{sec:experiments_fliplr}. Adding this feature allows us to shorten training to its final duration of 9.9 epochs, yielding our final training method \href{https://github.com/KellerJordan/cifar10-airbench/blob/04512094b7341d95ed02f697c5f5db404556137e/airbench94.py}{\texttt{airbench94.py}}, the entire contents of which can be found in \Secref{sec:code}. It reaches 94\% accuracy in 3.83 seconds on an NVIDIA A100.



\subsection{Compilation}
The final step we take to speed up training is a non-algorithmic one: we compile our training method using \texttt{torch.compile} in order to more efficiently utilize the GPU. This results in a training script which is mathematically equivalent (up to small differences in floating point arithmetic) to the non-compiled variant while being significantly faster: training time is reduced by 14\% to 3.29 A100-seconds. The downside is that the one-time compilation process takes up to several minutes to complete before training runs can begin, so that it is only beneficial when we plan to execute many runs of training at once. We release this version as \href{https://github.com/KellerJordan/cifar10-airbench/blob/04512094b7341d95ed02f697c5f5db404556137e/airbench94_compiled.py}{\texttt{airbench94\_compiled.py}}.

\section{95\% and 96\% targets}

To address scenarios where somewhat higher performance is desired, we additionally develop methods targeting 95\% and 96\% accuracy. Both are straightforward modifications \texttt{airbench94}.

To attain 95\% accuracy, we increase training epochs from 9.9 to 15, and we scale the output channel count of the first block from 64 to 128 and of the second two blocks from 256 to 384. We
reduce the learning rate by a factor of 0.87. These modifications yield \texttt{airbench95} which attains 95.01\% accuracy in 10.4 A100-seconds, consuming $1.4 \times 10^{15}$ FLOPs.

\begin{wrapfigure}{r}{0.45\textwidth}
\vspace{-5mm}
\begin{center}
    \includegraphics[width=0.44\textwidth]{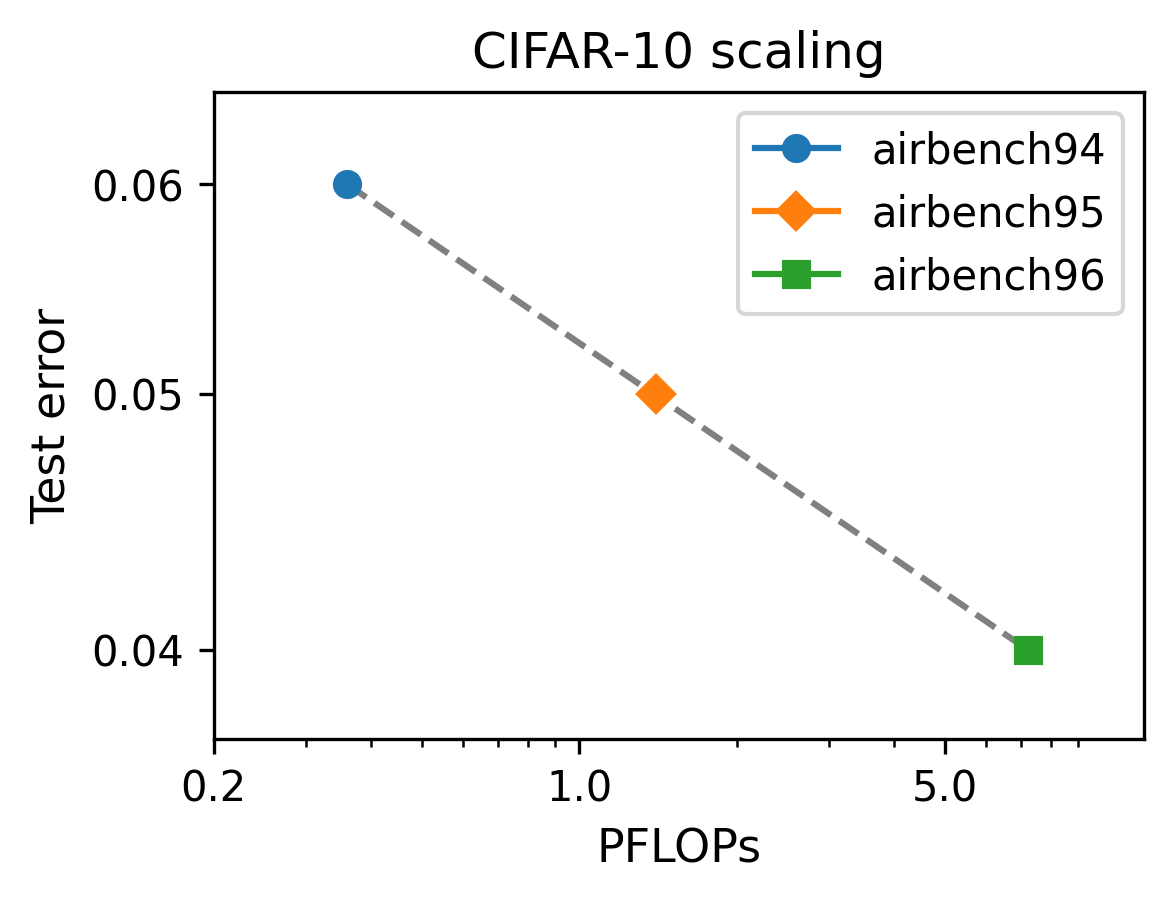}
    \vspace{-2mm}
    \caption{\small \textbf{FLOPs vs. error rate tradeoff.} Our three training methods apparently follow a linear log-log relationship between FLOPs and error rate.}
    \label{fig:scaling}
\end{center}
\vspace{-5mm}
\end{wrapfigure}

To attain 96\% accuracy, we add 12-pixel Cutout~\citep{devries2017improved} augmentation and raise the training epochs to 40. We add a third convolution to each block, and scale the first block to 128 channels and the second two to 512.
We also add a residual connection across the later two convolutions of each block, which we find is still beneficial despite the fact that we are already using identity initialization~(\Secref{sec:dirac}) to ease gradient flow.
Finally, we reduce the learning rate by a factor of 0.78. These changes yield \texttt{airbench96} which attains 96.05\% accuracy in 46.3 A100-seconds, consuming $7.2 \times 10^{15}$ FLOPs. \Figref{fig:scaling} shows the FLOPs and error rate of each of our three training methods.

\section{Experiments}
\label{sec:experiments}

\begin{figure}
    \centering
    \includegraphics[width=0.8\textwidth]{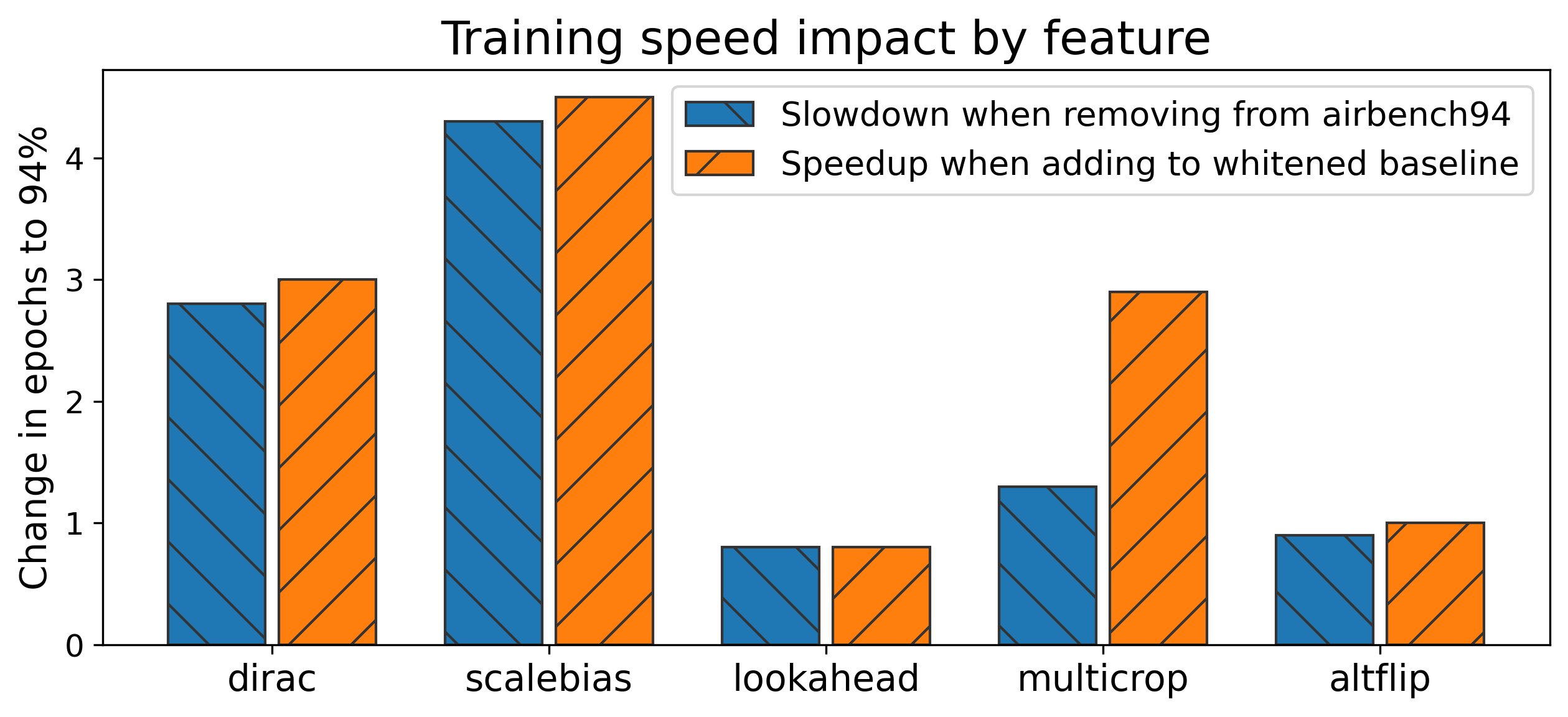}
    \caption{\small \textbf{Training speedups accumulate additively.} Removing individual features from \texttt{airbench94} increases the epochs-to-94\%. Adding the same features to the whitened baseline training~(\Secref{sec:whiten}) reduces the epochs-to-94\%. For every feature except multi-crop TTA~(\Secref{sec:multicrop}), these two changes in in epochs-to-94\% are roughly the same, suggesting that training speedups accumulate additively rather than multiplicatively.}
    \label{fig:additive}
\end{figure}

\subsection{Interaction between features}
To gain a better sense of the impact of each feature on training speed, we compare two quantities. First, we measure the number of epochs that can be saved by adding the feature to the whitened baseline~(\Secref{sec:whiten}). Second, we measure the number of epochs that must be added when the feature is removed from the final \texttt{airbench94}~(\Secref{sec:altflip}). For example, adding identity initialization~(\Secref{sec:dirac}) to the whitened baseline reduces the epochs-to-94\% from 21 to 18, and removing it from the final \texttt{airbench94} increases epochs-to-94\% from 9.9 to 12.8.

\Figref{fig:additive} shows both quantities for each feature. Surprisingly, we find that for all features except multi-crop TTA, the change in epochs attributable to a given feature is similar in both cases, even though the whitened baseline requires more than twice as many epochs as the final configuration. This indicates that the interaction between most features is additive rather than multiplicative.

\vspace{-2mm}
\subsection{Does alternating flip generalize?}
\vspace{-2mm}
\label{sec:experiments_fliplr}

\begin{figure}
    \centering
    \includegraphics[width=0.49\textwidth]{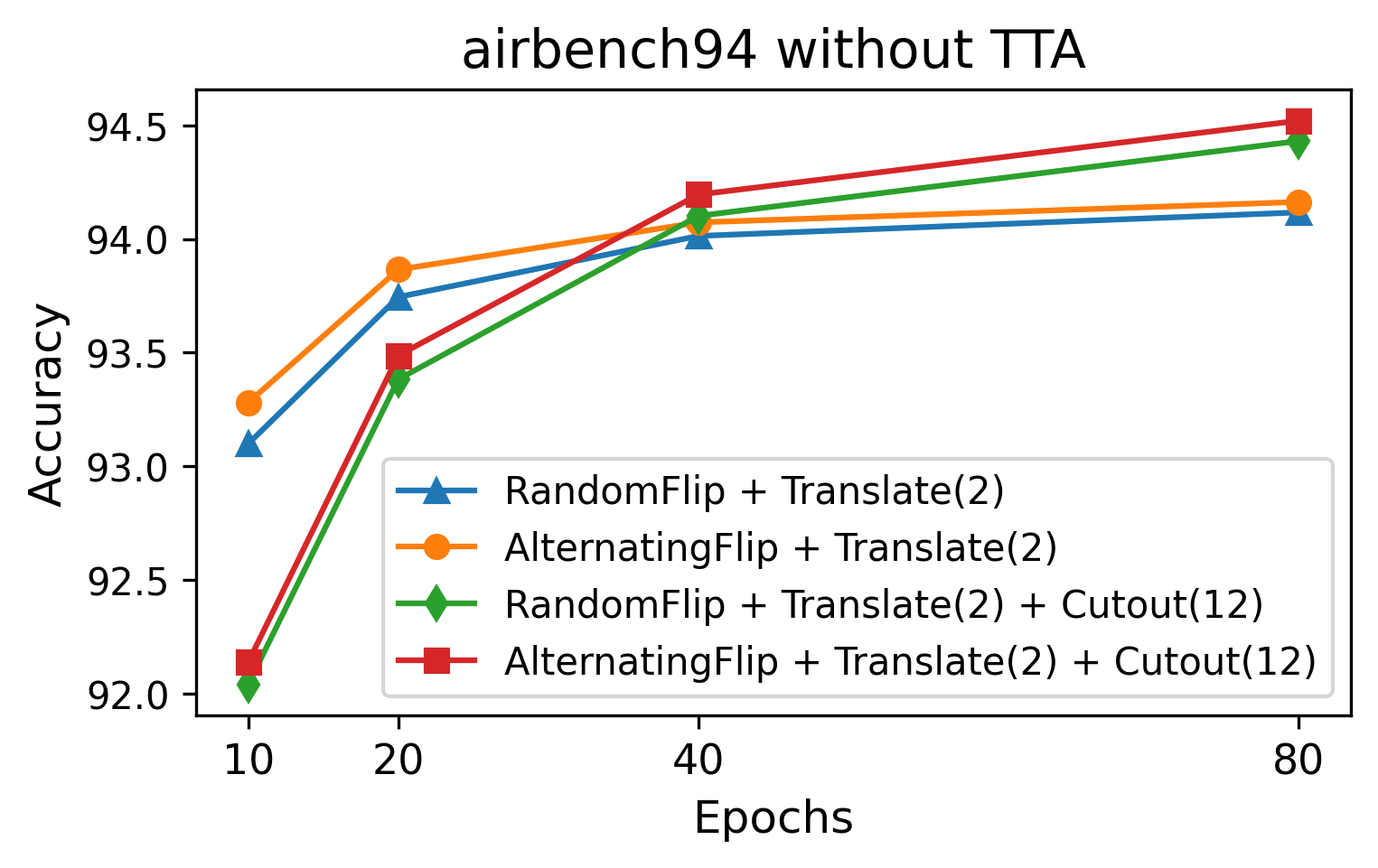}
    \includegraphics[width=0.49\textwidth]{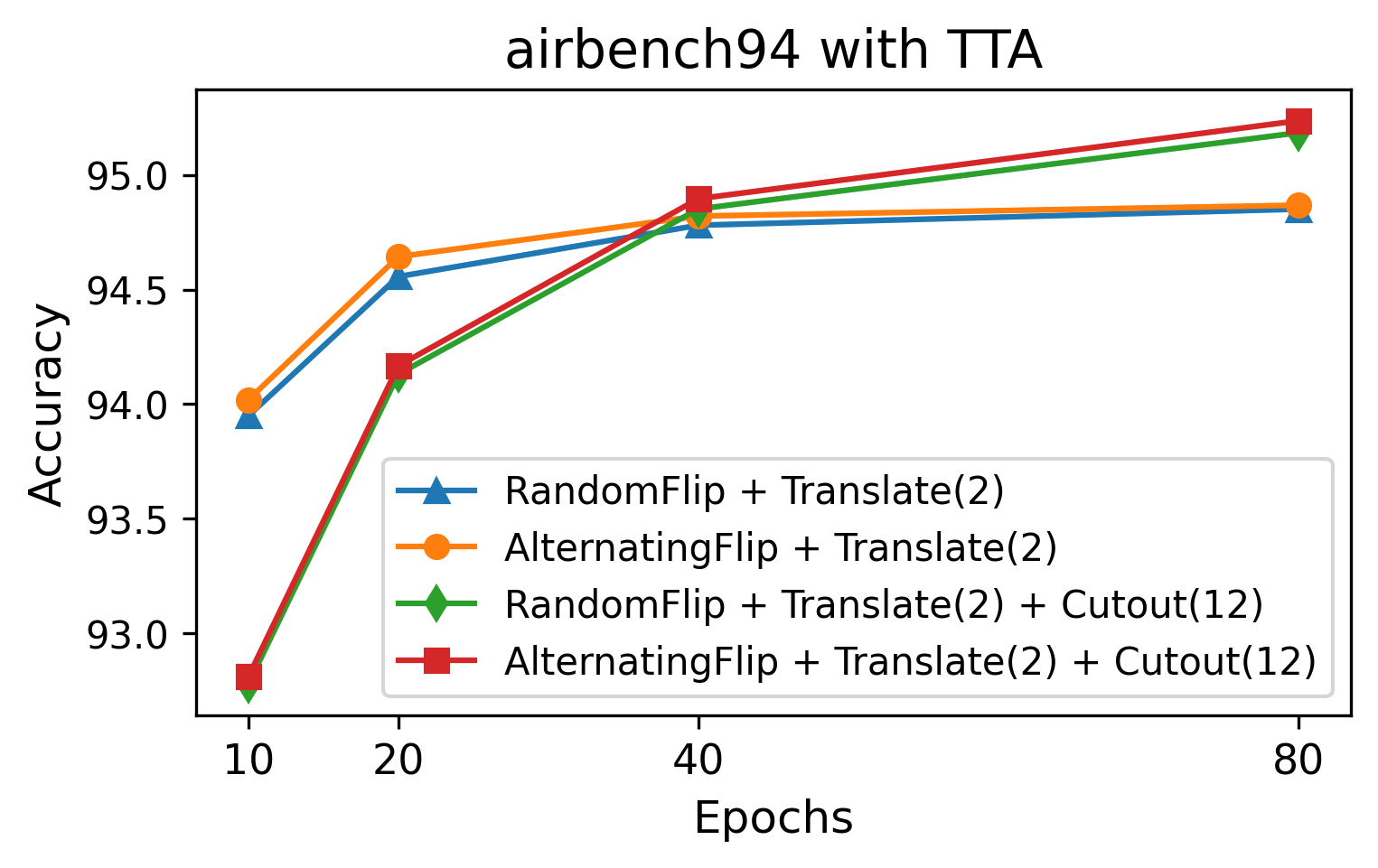}
    \includegraphics[width=0.49\textwidth]{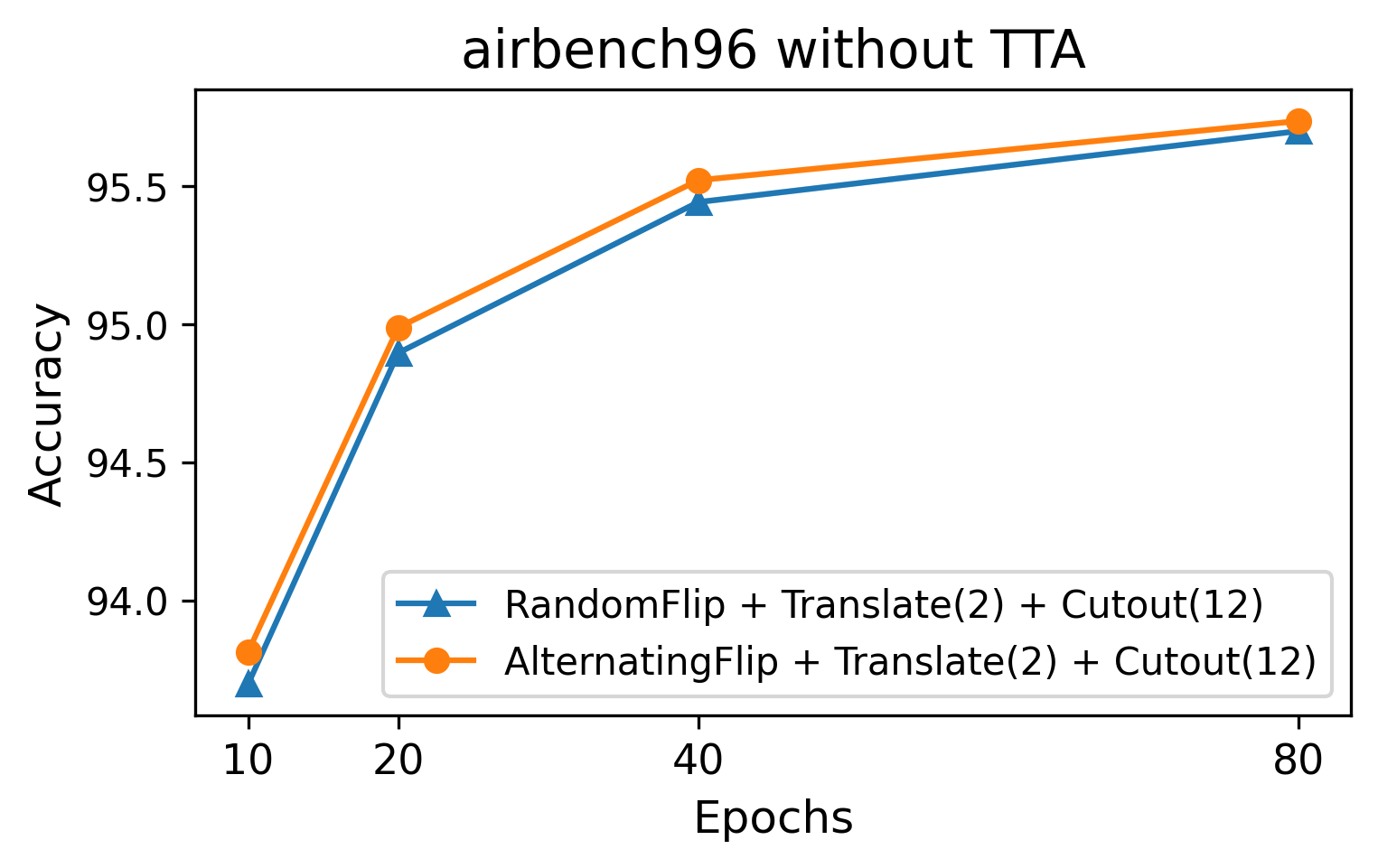}
    \includegraphics[width=0.49\textwidth]{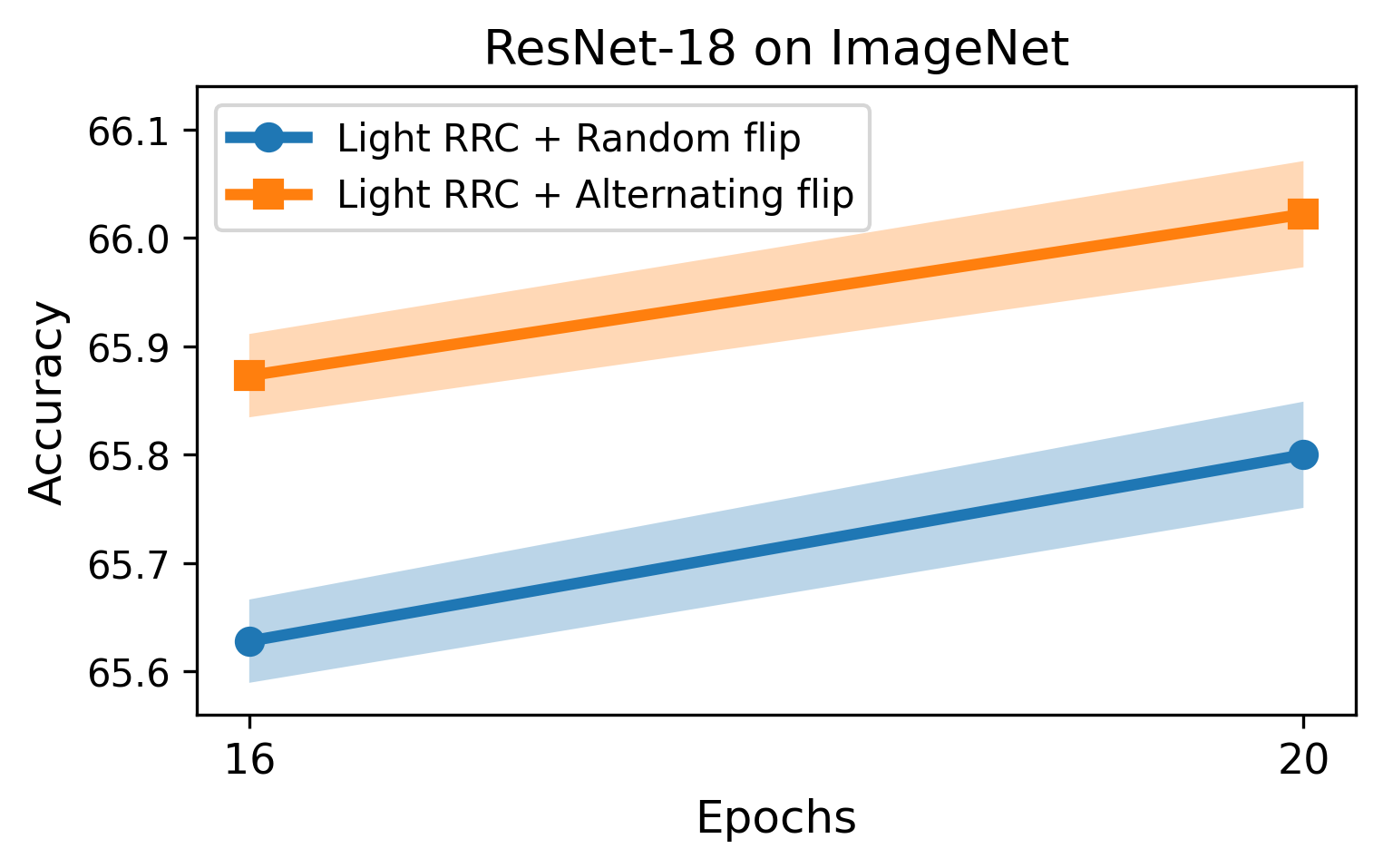}
    \vspace{-2mm}
    \caption{\small \textbf{Alternating flip boosts performance.} Across a variety of settings for \texttt{airbench94} and \texttt{airbench96}, the use of alternating flip rather than random flip consistently boosts performance by the equivalent of a 0-25\% training speedup. The benefit generalizes to ImageNet trainings which use light augmentation other than flipping.
    95\% confidence intervals are shown around each point.
    }
    \label{fig:altflip_evidence}
    \vspace{-3mm}
\end{figure}

In this section we investigate the effectiveness of alternating flip~(\Secref{sec:altflip}) across a variety of training configurations on CIFAR-10 and ImageNet. We find that it improves training speed in all cases except those where neither alternating nor random flip improve over using no flipping at all.

For CIFAR-10 we consider the performance boost given by alternating flip across the following 24 training configurations: \texttt{airbench94}, \texttt{airbench94} with extra Cutout augmentation, and \texttt{airbench96}, each with epochs in the range $\{10, 20, 40, 80\}$ and TTA~(\Secref{sec:multicrop}) in $\{\text{yes}, \text{no}\}$. For each configuration we compare the performance of alternating and random flip in terms of their mean accuracy across $n=400$ runs of training.

\Figref{fig:altflip_evidence} shows the result (see Table~\ref{tab:altflip_cifar10} for raw numbers). Switching from random flip to alternating flip improves performance in every setting. To get a sense for how big the improvement is, we estimate the effective speedup for each case, \ie the fraction of epochs that could be saved by switching from random to alternating flip while maintaining the level of accuracy of random flip.
We begin by fitting power law curves of the form $\mathrm{error} = c + b \cdot \mathrm{epochs}^a$ to the epochs-to-error curves of each random flip-based training configuration. We use these curves to calculate the effective speedup afforded by switching from random to alternating flip.
For example, \texttt{airbench94} with random flip and without TTA attains 6.26\% error when run for 20 epochs and 5.99\% when run for 40 epochs. The same configuration with alternating flip attains 6.13\% when run for 20 epochs, which a power-law fit predicts would take 25.3 epochs to attain using random flip. So we report a speedup of 27\%. Note that using a power-law yields a more conservative estimate relative to using linear interpolation between the observed epochs vs. error datapoints, which would yield a predicted speedup of 52\%.

Table~\ref{tab:altflip} shows the result. We observe the following patterns. First, the addition of extra augmentation (Cutout) somewhat closes the gap between random and alternating flip. To explain this, we note that the main effect of alternating flip is that it eliminates cases where an image is redundantly flipped the same way for many epochs in a row; we speculate that adding extra augmentation reduces the negative impact of these cases because it increases data diversity. Next, TTA reduces the gap between random and alternating flip. It also reduces the gap between random flip and no flipping at all~(Table~\ref{tab:altflip_cifar10}), indicating that TTA simply reduces the importance of flipping augmentation as such.
Finally, training for longer consistently increases the effective speedup given by alternating flip.

We next study ImageNet trainings with the following experiment. We train a ResNet-18 with a variety of train and test crops, comparing three flipping options: alternating flip, random flip, and no flipping at all. We consider two test crops: 256x256 center crop with crop ratio 0.875, and 192x192 center crop with crop ratio 1.0. We write CC(256, 0.875) to denote the former and CC(192, 1.0) to denote the latter. We also consider two training crops: 192x192 inception-style random resized crop~\citep{szegedy2014going}, which has aspect ratio ranging from 0.75 to 1.33 and covers an area ranging from 8\% to 100\% of the image, and a less aggressive random crop, which first resizes the shorter side of the image to 192 pixels, and then selects a random 192x192 square crop. We write Heavy RRC to denote the former and Light RRC to denote the latter. Full training details are provided in \Secref{sec:imagenet_details}.

Table~\ref{tab:imagenet} reports the mean top-1 validation accuracy of each case.
We first note that Heavy RRC is better when networks are evaluated with the CC(256, 0.875) crop, and Light RRC is slightly better when CC(192, 1.0) is used. This is fairly unsurprising given the standard theory of train-test resolution discrepancy~\citep{touvron2019fixing}.

For trainings which use Light RRC, we find that switching from random flip to alternating flip provides a substantial boost to performance, amounting to a training speedup of more than 25\%. In \Figref{fig:altflip_evidence} we visualize the improvement for short trainings with Light RRC, where switching to alternating flip improves performance by more than increasing the training duration from 16 to 20 epochs. The boost is higher when horizontal flipping TTA is turned off, which is consistent with our results on CIFAR-10. On the other hand, trainings which use Heavy RRC see no significant benefit from alternating flip. Indeed, even turning flipping off completely does not significantly reduce the performance of these trainings. We conclude that alternating flip improves over random flip for every training scenario where the latter improves over no flipping at all.



\begin{table}
\centering
\begin{tabular}{lll|ll}
    \toprule
    Baseline & Cutout & Epochs & Speedup & Speedup (w/ TTA) \\
    \midrule
    airbench94 & No & 10 & 15.0\% & \textit{5.30\%} \\
    airbench94 & No & 20 & 27.1\% & 21.3\% \\
    airbench94 & No & 40 & 38.3\% & 36.4\% \\
    airbench94 & No & 80 & 102\% & 31.8\% \\
    \midrule
    airbench94 & Yes & 10 & 3.84\% & 1.13\% \\
    airbench94 & Yes & 20 & 7.42\% & 2.00\% \\
    airbench94 & Yes & 40 & 18.6\% & 9.28\% \\
    airbench94 & Yes & 80 & 29.2\% & 14.25\% \\
    \midrule
    airbench96 & Yes & 10 & 4.94\% & 1.11\% \\
    airbench96 & Yes & 20 & 8.99\% & 3.58\% \\
    airbench96 & Yes & 40 & 17.2\% & \textit{6.48\%} \\
    airbench96 & Yes & 80 & 18.8\% & Not measured \\
    \bottomrule
\end{tabular}
\vspace{3mm}
\caption{\small Effective speedups given by switching from random flip to alternating flip. The two configurations most closely corresponding to \texttt{airbench94.py} and \texttt{airbench96.py} are italicized. See Table~\ref{tab:altflip_cifar10} for the raw accuracy values of the \texttt{airbench94} experiments.}
\label{tab:altflip}
\vspace{-3mm}
\end{table}

\begin{table}
\centering
\setlength{\tabcolsep}{5pt}
\begin{tabular}{llll|lll}
    \toprule
    &&&& \multicolumn{3}{c}{Flipping augmentation option} \\
    \cmidrule(r){5-7}
    Train crop & Test crop & Epochs & TTA & None & Random & Alternating \\
    \midrule
    Heavy RRC & CC(256, 0.875) & 16 & No & \textbf{66.78\%}$_{n=8}$ & 66.54\%$_{n=28}$ & 66.58\%$_{n=28}$ \\
    Heavy RRC & CC(192, 1.0) & 16 & No & 64.43\%$_{n=8}$ & 64.62\%$_{n=28}$ & 64.63\%$_{n=28}$ \\
    Light RRC & CC(256, 0.875) & 16 & No & 59.02\%$_{n=4}$ & 61.84\%$_{n=26}$ & \textbf{62.19\%}$_{n=26}$ \\
    Light RRC & CC(192, 1.0) & 16 & No & 61.79\%$_{n=4}$ & 64.50\%$_{n=26}$ & \textbf{64.93\%}$_{n=26}$ \\
    \midrule
    Heavy RRC & CC(256, 0.875) & 16 & Yes & 67.52\%$_{n=8}$ & 67.65\%$_{n=28}$ & 67.60\%$_{n=28}$ \\
    Heavy RRC & CC(192, 1.0) & 16 & Yes & 65.36\%$_{n=8}$ & 65.48\%$_{n=28}$ & 65.51\%$_{n=28}$ \\
    Light RRC & CC(256, 0.875) & 16 & Yes & 61.08\%$_{n=4}$ & 62.89\%$_{n=26}$ & \textbf{63.08\%}$_{n=26}$ \\
    Light RRC & CC(192, 1.0) & 16 & Yes & 63.91\%$_{n=4}$ & 65.63\%$_{n=26}$ & \textbf{65.87\%}$_{n=26}$ \\
    \midrule
    Light RRC & CC(192, 1.0) & 20 & Yes & not measured & 65.80\%$_{n=16}$ & \textbf{66.02\%}$_{n=16}$ \\
    \midrule
    Heavy RRC & CC(256, 0.875) & 88 & Yes & 72.34\%$_{n=2}$ & 72.45\%$_{n=4}$ & 72.46\%$_{n=4}$ \\
    \bottomrule
\end{tabular}
\vspace{3mm}
\caption{\small ImageNet validation accuracy for ResNet-18 trainings. Alternating flip improves over random flip for those trainings where random flip improves significantly over not flipping at all. The single best flipping option in each row is bolded when the difference is statistically significant.}
\label{tab:imagenet}
\vspace{-5mm}
\end{table}


\vspace{-1mm}
\subsection{Variance and class-wise calibration}
\label{sec:variance}
\vspace{-1mm}
Previous sections have focused on understanding what factors affect the first moment of accuracy (the mean). In this section we investigate the second moment, finding that TTA reduces variance at the cost of calibration.

Our experiment is to execute 10,000 runs of \texttt{airbench94} training with several hyperparameter settings. For each setting we report both the variance in test-set accuracy as well as an estimate of the distribution-wise variance~\citep{jordan2023calibrated}. \Figref{fig:distributions} shows the raw accuracy distributions.



Table~\ref{tab:variance} shows the results. Every case has at least $5\times$ less distribution-wise variance than test-set variance, replicating the main finding of \citet{jordan2023calibrated}. This is a surprising result because these trainings are at most 20 epochs, whereas the more standard training studied by \citet{jordan2023calibrated} had $5\times$ as much distribution-wise variance when run for a similar duration, and reached a low variance only when run for 64 epochs. We conclude from this comparison that distribution-wise variance is more strongly connected to the rate of convergence of a training rather than its duration as such. We also note that the low distribution-wise variance of \texttt{airbench94} indicates it has high training stability.

Using TTA significantly reduces the test-set variance, such that all three settings with TTA have lower test-set variance than any setting without TTA. However, test-set variance is implied by the class-wise calibration property~\citep{jordan2023calibrated,jiang2021assessing}, so contrapositively, we hypothesize that this reduction in test-set variance must come at the cost of class-wise calibration. To test this hypothesis, we compute the class-aggregated calibration error (CACE)~\citep{jiang2021assessing} of each setting, which measures deviation from class-wise calibration. Table~\ref{tab:variance} shows the results. Every setting with TTA has a higher CACE than every setting without TTA, confirming the hypothesis.

\begin{table}
  \centering
  \begin{tabular}{ccc|cccc}
    \toprule
    Epochs & Width & TTA & Mean accuracy & Test-set stddev & Dist-wise stddev & CACE \\
    \midrule
    $1\times$ & $1\times$ & No & 93.25\% & 0.157\% & 0.037\% & 0.0312 \\
    $2\times$ & $1\times$ & No & 93.86\% & 0.152\% & 0.025\% & 0.0233 \\
    $1.5\times$ & $1.5\times$ & No & 94.32\% & 0.142\% & 0.020\% & 0.0269 \\
    $1\times$ & $1\times$ & Yes & 94.01\% & 0.128\% & 0.029\% & 0.0533 \\
    $2\times$ & $1\times$ & Yes & 94.65\% & 0.124\% & 0.022\% & 0.0433 \\
    $1.5\times$ & $1.5\times$ & Yes & 94.97\% & 0.116\% & 0.018\% & 0.0444 \\
    \bottomrule
  \end{tabular}
  \vspace{3mm}
  \caption{Statistical metrics for \texttt{airbench94} trainings (n=10,000 runs each).}
  \label{tab:variance}
  \vspace{-5mm}
\end{table}


\vspace{-1mm}
\section{Discussion}
\vspace{-1mm}

In this paper we introduced a new training method for CIFAR-10. It reaches 94\% accuracy $1.9\times$ faster than the prior state-of-the-art, while being calibrated and highly stable. It is released as the \texttt{airbench} Python package.

We developed \texttt{airbench} solely with the goal of maximizing training speed on CIFAR-10. In \Secref{sec:generalization} we find that it also generalizes well to other tasks. For example, without any extra tuning, \texttt{airbench96} attains 1.7\% better performance than standard ResNet-18 when training on CIFAR-100.

One factor contributing to the training speed of \texttt{airbench} was our finding that
training can be accelerated by partially \textit{derandomizing} the standard random horizontal flipping augmentation, resulting in the variant that we call alternating flip~(\Figref{fig:viz_altflip}, \Secref{sec:altflip}).
Replacing random flip with alternating flip improves the performance of every training we considered~(\Secref{sec:experiments_fliplr}), with the exception of those trainings which do not benefit from horizontal flipping at all.
We note that, surprisingly to us, the standard ImageNet trainings that we considered do not significantly benefit from horizontal flipping.
Future work might investigate whether it is possible to obtain derandomized improvements to other augmentations besides horizontal flip.

The methods we introduced in this work improve the state-of-the-art for training speed on CIFAR-10, with fixed performance and hardware constraints.
These constraints mean that we cannot improve performance by simply scaling up the amount of computational resources used; instead we are forced to develop new methods like the alternating flip. We look forward to seeing what other new methods future work discovers to push training speed further.



\bibliography{references}
\bibliographystyle{iclr2024_conference}

\newpage
\appendix

\section{Network architecture}
\label{sec:arch}
\begin{lstlisting}[language=Python, caption=Minimal PyTorch code for the network architecture used by \texttt{airbench94}.]
from torch import nn

class Flatten(nn.Module):
    def forward(self, x): 
        return x.view(x.size(0), -1) 

class Mul(nn.Module):
    def __init__(self, scale):
        super().__init__()
        self.scale = scale
    def forward(self, x): 
        return x * self.scale

def conv(ch_in, ch_out):
    return nn.Conv2d(ch_in, ch_out, kernel_size=3,
                     padding='same', bias=False)

def make_net():
    act = lambda: nn.GELU()
    bn = lambda ch: nn.BatchNorm2d(ch)
    return nn.Sequential(
        nn.Sequential(
            nn.Conv2d(3, 24, kernel_size=2, padding=0, bias=True),
            act(),
        ),
        nn.Sequential(
            conv(24, 64),
            nn.MaxPool2d(2),
            bn(64), act(),
            conv(64, 64),
            bn(64), act(),
        ),
        nn.Sequential(
            conv(64, 256),
            nn.MaxPool2d(2),
            bn(256), act(),
            conv(256, 256),
            bn(256), act(),
        ),
        nn.Sequential(
            conv(256, 256),
            nn.MaxPool2d(2),
            bn(256), act(),
            conv(256, 256),
            bn(256), act(),
        ),
        nn.MaxPool2d(3),
        Flatten(),
        nn.Linear(256, 10, bias=False),
        Mul(1/9),
    )
\end{lstlisting}

We note that there exist various tweaks to the architecture which reduce FLOP usage but not wallclock time. For example, we can lower the FLOPs of \texttt{airbench96} by almost 20\% by reducing the kernel size of the first convolution in each block from 3 to 2 and increasing epochs from 40 to 45. But this does not improve the wallclock training time on an A100. Reducing the batch size is another easy way to save FLOPs but not wallclock time.

\section{Extra dataset experiments}
\label{sec:generalization}

\begin{table}
\centering
\setlength{\tabcolsep}{5pt}
\begin{tabular}{lll|ll}
    \toprule
    Dataset & Flipping? & Cutout? & ResNet-18 & \texttt{airbench96} \\
    \midrule
    CIFAR-10 & Yes & No & 95.55\% & 95.61\% \\
    CIFAR-10 & Yes & Yes & 96.01\% & 96.05\% \\
    CIFAR-100 & Yes & No & 77.54\% & 79.27\% \\
    CIFAR-100 & Yes & Yes & 78.04\% & 79.76\% \\
    CINIC-10 & Yes & No & 87.58\% & 87.78\% \\
    CINIC-10 & Yes & Yes & not measured & 88.22\% \\
    SVHN & No & No & 97.35\% & 97.38\% \\
    SVHN & No & Yes & not measured & 97.64\% \\
    \bottomrule
\end{tabular}
\vspace{3mm}
\caption{Comparison of \texttt{airbench96} to standard ResNet-18 training across a variety of tasks. We directly apply \texttt{airbench96} to each task without re-tuning any hyperparameters (besides turning off flipping for SVHN).}
\label{tab:compare_resnet18}
\end{table}

We developed \texttt{airbench} with the singular goal of maximizing training speed on CIFAR-10. To find out whether this has resulted in it being ``overfit'' to CIFAR-10, in this section we evaluate its performance on CIFAR-100~\citep{cifar100}, SVHN~\citep{netzer2011reading}, and CINIC-10~\citep{darlow2018cinic}.

On CIFAR-10, \texttt{airbench96} attains comparable accuracy to a standard ResNet-18 training, in both the case where both trainings use Cutout~\citep{devries2017improved} and the case where both do not (Table~\ref{tab:compare_resnet18}). So, if we evaluate \texttt{airbench96} on other tasks and find that it attains worse accuracy than ResNet-18, then we can say that \texttt{airbench96} must be overfit to CIFAR-10, otherwise we can say that it generalizes.

We compare to the best accuracy numbers we can find in the literature for ResNet-18 on each task. We do not tune the hyperparameters of \texttt{airbench96} at all: we use the same values that were optimal on CIFAR-10. Table~\ref{tab:compare_resnet18} shows the result. It turns out that in every case, \texttt{airbench96} attains better performance than ResNet-18 training. Particularly impressive are results on CIFAR-100 where \texttt{airbench96} attains 1.7\% higher accuracy than ResNet-18 training, both in the case that Cutout is used and the case that it is not. We conclude that \texttt{airbench} is not overfit to CIFAR-10, since it shows strong generalization to other tasks.

We note that this comparison between \texttt{airbench96} and ResNet-18 training is fair in the sense that it does demonstrate that the former has good generalization, but unfair in the sense that it does not indicate that \texttt{airbench96} is the superior training as such. In particular, \texttt{airbench96} uses test-time augmentation whereas standard ResNet-18 training does not. It is likely that ResNet-18 training would outperform \texttt{airbench96} if it were run using test-time augmentation. However, it also takes 5-10 times longer to complete. The decision of which to use may be situational.

The accuracy values we report for ResNet-18 training are from the following sources. We tried to select the highest values we could find for each setting. \citet{moreau2022benchopt} reports attaining 95.55\% on CIFAR-10 without Cutout, and 97.35\% on SVHN. \citet{devries2017improved} reports attaining 96.01\% on CIFAR-10 with Cutout, 77.54\% on CIFAR-100 without Cutout, and 78.04\% on CIFAR-100 with Cutout. \citet{darlow2018cinic} report attaining 87.58\% on CINIC-10 without Cutout.




\section{ImageNet training details}
\label{sec:imagenet_details}
Our ImageNet trainings follow the 16 and 88-epoch configurations from \url{https://github.com/libffcv/ffcv-imagenet}. In particular, we use a batch size of 1024 and learning rate 0.5 and momentum 0.9, with a linear warmup and decay schedule for the learning rate. We train at resolution 160 for the majority of training and then ramp up to resolution 192 for roughly the last 30\% of training. We use label smoothing of 0.1. We use the FFCV~\citep{leclerc2023ffcv} data loader.

\section{Extra tables \& figures}


\begin{figure}[H]
    \centering
    \includegraphics[width=0.95\textwidth]{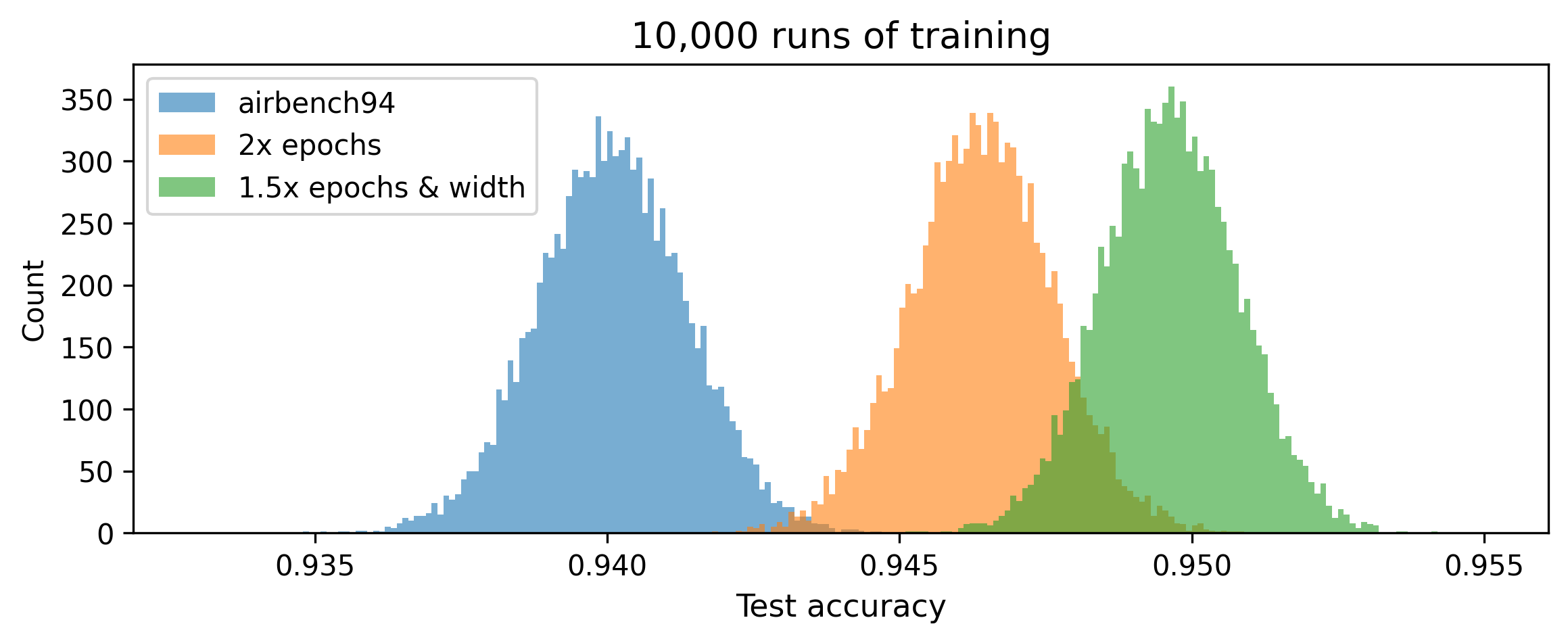}
    \caption{Accuracy distributions for the three \texttt{airbench94} variations (with TTA) described in \Secref{sec:variance}.}
    \label{fig:distributions}
\end{figure}

\begin{table}
\centering
\setlength{\tabcolsep}{5pt}
\begin{tabular}{lll|lll}
    \toprule
    \multicolumn{3}{c}{Hyperparameters} & \multicolumn{3}{c}{Flipping augmentation option} \\
    \cmidrule(r){1-3}\cmidrule(r){4-6}
    Epochs & Cutout & TTA & None & Random & Alternating \\
    \midrule
    10 & No & No & 92.3053 & 93.0988 & \textbf{93.2798} \\
    20 & No & No & 92.8166 & 93.7446 & \textbf{93.8652} \\
    40 & No & No & 93.0143 & 94.0133 & \textbf{94.0729} \\
    80 & No & No & 93.0612 & 94.1169 & \textbf{94.1628} \\
    \midrule
    10 & No & Yes & 93.4071 & 93.9488 & \textbf{94.0186} \\
    20 & No & Yes & 93.8528 & 94.5565 & \textbf{94.6530} \\
    40 & No & Yes & 94.0381 & 94.7803 & \textbf{94.8203} \\
    80 & No & Yes & 94.0638 & 94.8506 & \textbf{94.8676} \\
    \midrule
    10 & Yes & No & 91.8487 & 92.0402 & \textbf{92.1374} \\
    20 & Yes & No & 92.8474 & 93.3825 & \textbf{93.4876} \\
    40 & Yes & No & 93.2675 & 94.1014 & \textbf{94.1952} \\
    80 & Yes & No & 93.4193 & 94.4311 & \textbf{94.5204} \\
    \midrule
    10 & Yes & Yes & 92.6455 & 92.7780 & \textbf{92.8103} \\
    20 & Yes & Yes & 93.7862 & 94.1306 & \textbf{94.1670} \\
    40 & Yes & Yes & 94.3090 & 94.8511 & \textbf{94.8960} \\
    80 & Yes & Yes & 94.5253 & 95.1839 & \textbf{95.2362} \\
    \bottomrule
\end{tabular}
\vspace{3mm}
\caption{Raw accuracy values for \texttt{airbench94} flipping augmentation experiments. Each value is a mean over $n=400$ runs. The 95\% confidence intervals are roughly $\pm 0.014$, so that every row-wise difference in means is statistically significant.
}
\label{tab:altflip_cifar10}
\end{table}

\section{Complete training code}
\label{sec:code}
\begin{lstlisting}[language=Python, caption=\texttt{airbench94.py}]
"""
airbench94.py
3.83s runtime on an A100; 0.36 PFLOPs.
Evidence for validity: 94.01 average accuracy in n=1000 runs.

We recorded the runtime of 3.83 seconds on an NVIDIA A100-SXM4-80GB with the following nvidia-smi:
NVIDIA-SMI 515.105.01   Driver Version: 515.105.01   CUDA Version: 11.7
torch.__version__ == '2.1.2+cu118'
"""

#############################################
#            Setup/Hyperparameters          #
#############################################

import os
import sys
import uuid
from math import ceil

import torch
from torch import nn
import torch.nn.functional as F
import torchvision
import torchvision.transforms as T

torch.backends.cudnn.benchmark = True

"""
We express the main training hyperparameters (batch size, learning rate, momentum, and weight decay) in decoupled form, so that each one can be tuned independently. This accomplishes the following:
* Assuming time-constant gradients, the average step size is decoupled from everything but the lr.
* The size of the weight decay update is decoupled from everything but the wd.
In constrast, normally when we increase the (Nesterov) momentum, this also scales up the step size proportionally to 1 + 1 / (1 - momentum), meaning we cannot change momentum without having to re-tune the learning rate. Similarly, normally when we increase the learning rate this also increases the size of the weight decay, requiring a proportional decrease in the wd to maintain the same decay strength.

The practical impact is that hyperparameter tuning is faster, since this parametrization allows each one to be tuned independently. See https://myrtle.ai/learn/how-to-train-your-resnet-5-hyperparameters/.
"""

hyp = {
    'opt': {
        'train_epochs': 9.9,
        'batch_size': 1024,
        'lr': 11.5,                 # learning rate per 1024 examples
        'momentum': 0.85,
        'weight_decay': 0.0153,     # weight decay per 1024 examples (decoupled from learning rate)
        'bias_scaler': 64.0,        # scales up learning rate (but not weight decay) for BatchNorm biases
        'label_smoothing': 0.2,
        'whiten_bias_epochs': 3,    # how many epochs to train the whitening layer bias before freezing
    },
    'aug': {
        'flip': True,
        'translate': 2,
    },
    'net': {
        'widths': {
            'block1': 64,
            'block2': 256,
            'block3': 256,
        },
        'batchnorm_momentum': 0.6,
        'scaling_factor': 1/9,
        'tta_level': 2,         # the level of test-time augmentation: 0=none, 1=mirror, 2=mirror+translate
    },
}

#############################################
#                DataLoader                 #
#############################################

CIFAR_MEAN = torch.tensor((0.4914, 0.4822, 0.4465))
CIFAR_STD = torch.tensor((0.2470, 0.2435, 0.2616))

def batch_flip_lr(inputs):
    flip_mask = (torch.rand(len(inputs), device=inputs.device) < 0.5).view(-1, 1, 1, 1)
    return torch.where(flip_mask, inputs.flip(-1), inputs)

def batch_crop(images, crop_size):
    r = (images.size(-1) - crop_size)//2
    shifts = torch.randint(-r, r+1, size=(len(images), 2), device=images.device)
    images_out = torch.empty((len(images), 3, crop_size, crop_size), device=images.device, dtype=images.dtype)
    # The two cropping methods in this if-else produce equivalent results, but the second is faster for r > 2.
    if r <= 2:
        for sy in range(-r, r+1):
            for sx in range(-r, r+1):
                mask = (shifts[:, 0] == sy) & (shifts[:, 1] == sx)
                images_out[mask] = images[mask, :, r+sy:r+sy+crop_size, r+sx:r+sx+crop_size]
    else:
        images_tmp = torch.empty((len(images), 3, crop_size, crop_size+2*r), device=images.device, dtype=images.dtype)
        for s in range(-r, r+1):
            mask = (shifts[:, 0] == s)
            images_tmp[mask] = images[mask, :, r+s:r+s+crop_size, :]
        for s in range(-r, r+1):
            mask = (shifts[:, 1] == s)
            images_out[mask] = images_tmp[mask, :, :, r+s:r+s+crop_size]
    return images_out

class CifarLoader:
    """
    GPU-accelerated dataloader for CIFAR-10 which implements alternating flip augmentation.
    """

    def __init__(self, path, train=True, batch_size=500, aug=None, drop_last=None, shuffle=None, gpu=0):
        data_path = os.path.join(path, 'train.pt' if train else 'test.pt')
        if not os.path.exists(data_path):
            dset = torchvision.datasets.CIFAR10(path, download=True, train=train)
            images = torch.tensor(dset.data)
            labels = torch.tensor(dset.targets)
            torch.save({'images': images, 'labels': labels, 'classes': dset.classes}, data_path)

        data = torch.load(data_path, map_location=torch.device(gpu))
        self.images, self.labels, self.classes = data['images'], data['labels'], data['classes']
        # It's faster to load+process uint8 data than to load preprocessed fp16 data
        self.images = (self.images.half() / 255).permute(0, 3, 1, 2).to(memory_format=torch.channels_last)

        self.normalize = T.Normalize(CIFAR_MEAN, CIFAR_STD)
        self.proc_images = {} # Saved results of image processing to be done on the first epoch
        self.epoch = 0

        self.aug = aug or {}
        for k in self.aug.keys():
            assert k in ['flip', 'translate'], 'Unrecognized key: %s' % k

        self.batch_size = batch_size
        self.drop_last = train if drop_last is None else drop_last
        self.shuffle = train if shuffle is None else shuffle

    def __len__(self):
        return len(self.images)//self.batch_size if self.drop_last else ceil(len(self.images)/self.batch_size)

    def __iter__(self):

        if self.epoch == 0:
            images = self.proc_images['norm'] = self.normalize(self.images)
            # Randomly flip all images on the first epoch as according to definition of alternating flip
            if self.aug.get('flip', False):
                images = self.proc_images['flip'] = batch_flip_lr(images)
            # Pre-pad images to save time when doing random translation
            pad = self.aug.get('translate', 0)
            if pad > 0:
                self.proc_images['pad'] = F.pad(images, (pad,)*4, 'reflect')

        if self.aug.get('translate', 0) > 0:
            images = batch_crop(self.proc_images['pad'], self.images.shape[-2])
        elif self.aug.get('flip', False):
            images = self.proc_images['flip']
        else:
            images = self.proc_images['norm']
        if self.aug.get('flip', False):
            if self.epoch % 2 == 1:
                images = images.flip(-1)

        self.epoch += 1

        indices = (torch.randperm if self.shuffle else torch.arange)(len(images), device=images.device)
        for i in range(len(self)):
            idxs = indices[i*self.batch_size:(i+1)*self.batch_size]
            yield (images[idxs], self.labels[idxs])

#############################################
#            Network Components             #
#############################################

class Flatten(nn.Module):
    def forward(self, x):
        return x.view(x.size(0), -1)

class Mul(nn.Module):
    def __init__(self, scale):
        super().__init__()
        self.scale = scale
    def forward(self, x):
        return x * self.scale

class BatchNorm(nn.BatchNorm2d):
    def __init__(self, num_features, momentum, eps=1e-12,
                 weight=False, bias=True):
        super().__init__(num_features, eps=eps, momentum=1-momentum)
        self.weight.requires_grad = weight
        self.bias.requires_grad = bias
        # Note that PyTorch already initializes the weights to one and biases to zero

class Conv(nn.Conv2d):
    def __init__(self, in_channels, out_channels, kernel_size=3, padding='same', bias=False):
        super().__init__(in_channels, out_channels, kernel_size=kernel_size, padding=padding, bias=bias)

    def reset_parameters(self):
        super().reset_parameters()
        if self.bias is not None:
            self.bias.data.zero_()
        w = self.weight.data
        torch.nn.init.dirac_(w[:w.size(1)])

class ConvGroup(nn.Module):
    def __init__(self, channels_in, channels_out, batchnorm_momentum):
        super().__init__()
        self.conv1 = Conv(channels_in,  channels_out)
        self.pool = nn.MaxPool2d(2)
        self.norm1 = BatchNorm(channels_out, batchnorm_momentum)
        self.conv2 = Conv(channels_out, channels_out)
        self.norm2 = BatchNorm(channels_out, batchnorm_momentum)
        self.activ = nn.GELU()

    def forward(self, x):
        x = self.conv1(x)
        x = self.pool(x)
        x = self.norm1(x)
        x = self.activ(x)
        x = self.conv2(x)
        x = self.norm2(x)
        x = self.activ(x)
        return x

#############################################
#            Network Definition             #
#############################################

def make_net(widths=hyp['net']['widths'], batchnorm_momentum=hyp['net']['batchnorm_momentum']):
    whiten_kernel_size = 2
    whiten_width = 2 * 3 * whiten_kernel_size**2
    net = nn.Sequential(
        Conv(3, whiten_width, whiten_kernel_size, padding=0, bias=True),
        nn.GELU(),
        ConvGroup(whiten_width,     widths['block1'], batchnorm_momentum),
        ConvGroup(widths['block1'], widths['block2'], batchnorm_momentum),
        ConvGroup(widths['block2'], widths['block3'], batchnorm_momentum),
        nn.MaxPool2d(3),
        Flatten(),
        nn.Linear(widths['block3'], 10, bias=False),
        Mul(hyp['net']['scaling_factor']),
    )
    net[0].weight.requires_grad = False
    net = net.half().cuda()
    net = net.to(memory_format=torch.channels_last)
    for mod in net.modules():
        if isinstance(mod, BatchNorm):
            mod.float()
    return net

#############################################
#       Whitening Conv Initialization       #
#############################################

def get_patches(x, patch_shape):
    c, (h, w) = x.shape[1], patch_shape
    return x.unfold(2,h,1).unfold(3,w,1).transpose(1,3).reshape(-1,c,h,w).float()

def get_whitening_parameters(patches):
    n,c,h,w = patches.shape
    patches_flat = patches.view(n, -1)
    est_patch_covariance = (patches_flat.T @ patches_flat) / n
    eigenvalues, eigenvectors = torch.linalg.eigh(est_patch_covariance, UPLO='U')
    return eigenvalues.flip(0).view(-1, 1, 1, 1), eigenvectors.T.reshape(c*h*w,c,h,w).flip(0)

def init_whitening_conv(layer, train_set, eps=5e-4):
    patches = get_patches(train_set, patch_shape=layer.weight.data.shape[2:])
    eigenvalues, eigenvectors = get_whitening_parameters(patches)
    eigenvectors_scaled = eigenvectors / torch.sqrt(eigenvalues + eps)
    layer.weight.data[:] = torch.cat((eigenvectors_scaled, -eigenvectors_scaled))

############################################
#                Lookahead                 #
############################################

class LookaheadState:
    def __init__(self, net):
        self.net_ema = {k: v.clone() for k, v in net.state_dict().items()}

    def update(self, net, decay):
        for ema_param, net_param in zip(self.net_ema.values(), net.state_dict().values()):
            if net_param.dtype in (torch.half, torch.float):
                ema_param.lerp_(net_param, 1-decay)
                net_param.copy_(ema_param)

############################################
#                 Logging                  #
############################################

def print_columns(columns_list, is_head=False, is_final_entry=False):
    print_string = ''
    for col in columns_list:
        print_string += '|  %s  ' % col
    print_string += '|'
    if is_head:
        print('-'*len(print_string))
    print(print_string)
    if is_head or is_final_entry:
        print('-'*len(print_string))

logging_columns_list = ['run   ', 'epoch', 'train_loss', 'train_acc', 'val_acc', 'tta_val_acc', 'total_time_seconds']
def print_training_details(variables, is_final_entry):
    formatted = []
    for col in logging_columns_list:
        var = variables.get(col.strip(), None)
        if type(var) in (int, str):
            res = str(var)
        elif type(var) is float:
            res = '{:0.4f}'.format(var)
        else:
            assert var is None
            res = ''
        formatted.append(res.rjust(len(col)))
    print_columns(formatted, is_final_entry=is_final_entry)

############################################
#               Evaluation                 #
############################################

def infer(model, loader, tta_level=0):
    """
    Test-time augmentation strategy (for tta_level=2):
    1. Flip/mirror the image left-to-right (50% of the time).
    2. Translate the image by one pixel either up-and-left or down-and-right (50% of the time, i.e. both happen 25% of the time).

    This creates 6 views per image (left/right times the two translations and no-translation), which we evaluate and then weight according to the given probabilities.
    """

    def infer_basic(inputs, net):
        return net(inputs).clone()

    def infer_mirror(inputs, net):
        return 0.5 * net(inputs) + 0.5 * net(inputs.flip(-1))

    def infer_mirror_translate(inputs, net):
        logits = infer_mirror(inputs, net)
        pad = 1
        padded_inputs = F.pad(inputs, (pad,)*4, 'reflect')
        inputs_translate_list = [
            padded_inputs[:, :, 0:32, 0:32],
            padded_inputs[:, :, 2:34, 2:34],
        ]
        logits_translate_list = [infer_mirror(inputs_translate, net)
                                 for inputs_translate in inputs_translate_list]
        logits_translate = torch.stack(logits_translate_list).mean(0)
        return 0.5 * logits + 0.5 * logits_translate

    model.eval()
    test_images = loader.normalize(loader.images)
    infer_fn = [infer_basic, infer_mirror, infer_mirror_translate][tta_level]
    with torch.no_grad():
        return torch.cat([infer_fn(inputs, model) for inputs in test_images.split(2000)])

def evaluate(model, loader, tta_level=0):
    logits = infer(model, loader, tta_level)
    return (logits.argmax(1) == loader.labels).float().mean().item()

############################################
#                Training                  #
############################################

def main(run):

    batch_size = hyp['opt']['batch_size']
    epochs = hyp['opt']['train_epochs']
    momentum = hyp['opt']['momentum']
    # Assuming gradients are constant in time, for Nesterov momentum, the below ratio is how much larger the default steps will be than the underlying per-example gradients. We divide the learning rate by this ratio in order to ensure steps are the same scale as gradients, regardless of the choice of momentum.
    kilostep_scale = 1024 * (1 + 1 / (1 - momentum))
    lr = hyp['opt']['lr'] / kilostep_scale # un-decoupled learning rate for PyTorch SGD
    wd = hyp['opt']['weight_decay'] * batch_size / kilostep_scale
    lr_biases = lr * hyp['opt']['bias_scaler']

    loss_fn = nn.CrossEntropyLoss(label_smoothing=hyp['opt']['label_smoothing'], reduction='none')
    test_loader = CifarLoader('cifar10', train=False, batch_size=2000)
    train_loader = CifarLoader('cifar10', train=True, batch_size=batch_size, aug=hyp['aug'])
    if run == 'warmup':
        # The only purpose of the first run is to warmup, so we can use dummy data
        train_loader.labels = torch.randint(0, 10, size=(len(train_loader.labels),), device=train_loader.labels.device)
    total_train_steps = ceil(len(train_loader) * epochs)

    model = make_net()
    current_steps = 0

    norm_biases = [p for k, p in model.named_parameters() if 'norm' in k and p.requires_grad]
    other_params = [p for k, p in model.named_parameters() if 'norm' not in k and p.requires_grad]
    param_configs = [dict(params=norm_biases, lr=lr_biases, weight_decay=wd/lr_biases),
                     dict(params=other_params, lr=lr, weight_decay=wd/lr)]
    optimizer = torch.optim.SGD(param_configs, momentum=momentum, nesterov=True)

    def triangle(steps, start=0, end=0, peak=0.5):
        xp = torch.tensor([0, int(peak * steps), steps])
        fp = torch.tensor([start, 1, end])
        x = torch.arange(1+steps)
        m = (fp[1:] - fp[:-1]) / (xp[1:] - xp[:-1])
        b = fp[:-1] - (m * xp[:-1])
        indices = torch.sum(torch.ge(x[:, None], xp[None, :]), 1) - 1
        indices = torch.clamp(indices, 0, len(m) - 1)
        return m[indices] * x + b[indices]
    lr_schedule = triangle(total_train_steps, start=0.2, end=0.07, peak=0.23)
    scheduler = torch.optim.lr_scheduler.LambdaLR(optimizer, lambda i: lr_schedule[i])

    alpha_schedule = 0.95**5 * (torch.arange(total_train_steps+1) / total_train_steps)**3
    lookahead_state = LookaheadState(model)

    # For accurately timing GPU code
    starter = torch.cuda.Event(enable_timing=True)
    ender = torch.cuda.Event(enable_timing=True)
    total_time_seconds = 0.0

    # Initialize the first layer using statistics of training images
    starter.record()
    train_images = train_loader.normalize(train_loader.images[:5000])
    init_whitening_conv(model[0], train_images)
    ender.record()
    torch.cuda.synchronize()
    total_time_seconds += 1e-3 * starter.elapsed_time(ender)

    for epoch in range(ceil(epochs)):

        model[0].bias.requires_grad = (epoch < hyp['opt']['whiten_bias_epochs'])

        ####################
        #     Training     #
        ####################

        starter.record()

        model.train()
        for inputs, labels in train_loader:

            outputs = model(inputs)
            loss = loss_fn(outputs, labels).sum()
            optimizer.zero_grad(set_to_none=True)
            loss.backward()
            optimizer.step()
            scheduler.step()

            current_steps += 1

            if current_steps % 5 == 0:
                lookahead_state.update(model, decay=alpha_schedule[current_steps].item())

            if current_steps >= total_train_steps:
                if lookahead_state is not None:
                    lookahead_state.update(model, decay=1.0)
                break

        ender.record()
        torch.cuda.synchronize()
        total_time_seconds += 1e-3 * starter.elapsed_time(ender)

        ####################
        #    Evaluation    #
        ####################

        # Print the accuracy and loss from the last training batch of the epoch
        train_acc = (outputs.detach().argmax(1) == labels).float().mean().item()
        train_loss = loss.item() / batch_size
        val_acc = evaluate(model, test_loader, tta_level=0)
        print_training_details(locals(), is_final_entry=False)
        run = None # Only print the run number once

    ####################
    #  TTA Evaluation  #
    ####################

    starter.record()
    tta_val_acc = evaluate(model, test_loader, tta_level=hyp['net']['tta_level'])
    ender.record()
    torch.cuda.synchronize()
    total_time_seconds += 1e-3 * starter.elapsed_time(ender)

    epoch = 'eval'
    print_training_details(locals(), is_final_entry=True)

    return tta_val_acc

if __name__ == "__main__":
    with open(sys.argv[0]) as f:
        code = f.read()

    print_columns(logging_columns_list, is_head=True)
    main('warmup')
    accs = torch.tensor([main(run) for run in range(25)])
    print('Mean: %.4f    Std: %.4f' % (accs.mean(), accs.std()))

    log = {'code': code, 'accs': accs}
    log_dir = os.path.join('logs', str(uuid.uuid4()))
    os.makedirs(log_dir, exist_ok=True)
    log_path = os.path.join(log_dir, 'log.pt')
    print(os.path.abspath(log_path))
    torch.save(log, os.path.join(log_dir, 'log.pt'))
\end{lstlisting}

\end{document}